\documentclass{article}

\usepackage{arxiv}

\usepackage[utf8]{inputenc} 
\usepackage[T1]{fontenc}    
\usepackage{hyperref}       
\usepackage{url}            
\usepackage{booktabs}       
\usepackage{amsfonts}       
\usepackage{nicefrac}       
\usepackage{microtype}      
\usepackage{graphicx}
\usepackage[sort,numbers]{natbib}
\usepackage{doi}

\usepackage{amsmath,amssymb}
\usepackage{algorithmic}
\usepackage{textcomp}

\usepackage{subfigure}

\usepackage[dvipsnames]{xcolor}
\definecolor{lightergray}{rgb}{0.8, 0.8, 0.8}
\usepackage{authblk}
\usepackage{multicol}
\usepackage{placeins}

\makeatletter
\newcommand{\printfnsymbol}[2]{%
  \textsuperscript{\@fnsymbol{#2}}%
}

\title{Highlighting the Importance of Reducing Research Bias and Carbon Emissions in CNNs}

\author{\textbf{Ahmed Badar, Arnav Varma, Adrian Staniec, Mahmoud Gamal, Omar Magdy, Haris Iqbal, Elahe Arani\thanks{Equal Advising} and Bahram Zonooz$^*$} \\ 
{Advanced Research Lab, Navinfo Europe, Eindhoven, The Netherlands}}

\date{}

\renewcommand{\headeright}{}
\renewcommand{\undertitle}{}

\hypersetup{
pdftitle={Highlighting the Importance of Reducing Research Bias and Carbon Emissions in CNNs},
pdfsubject={Deep Learning},
pdfauthor={Ahmed Badar, Arnav Varma, Adrian Staniec, Mahmoud Gamal, Omar Magdy, Haris Iqbal, Elahe Arani and Bahram Zonooz},
pdfkeywords={Green AI, Carbon emmissions, Image Segmentation and Classification},
}

\begin{document}

\maketitle
\begin{abstract}
Convolutional neural networks (CNNs) have become commonplace in addressing major challenges in computer vision. 
Researchers are not only coming up with new CNN architectures but are also researching different techniques to improve the performance of existing architectures.
However, there is a tendency to over-emphasize performance improvement while neglecting certain important variables such as simplicity, versatility, the fairness of comparisons, and energy efficiency. 
Overlooking these variables in architectural design and evaluation has led to research bias and a significantly negative environmental impact. Furthermore, this can undermine the positive impact of research in using deep learning models to tackle climate change.
Here, we perform an extensive and fair empirical study of a number of proposed techniques to gauge the utility of each technique for segmentation and classification. Our findings restate the importance of favoring simplicity over complexity in model design (Occam's Razor). Furthermore, our results indicate that simple standardized practices can lead to a significant reduction in environmental impact with little drop in performance. We highlight that there is a need to rethink the design and evaluation of CNNs to alleviate the issue of research bias and carbon emissions.
\end{abstract}


\section{Introduction}
\label{sec:introduction}
Deep neural networks (DNNs) have achieved remarkable results in recent years~\citep{lecun_deep_2015}, in applications such as image classification~\citep{he2016deep, krizhevsky_imagenet_2012}, speech recognition~\citep{deng2013new} and automation~\citep{bojarski2016end}.
A number of techniques have been proposed to further improve the performance of the networks which target different aspects of the learning and inference process.
This has led to development of techniques including different learning rate schedulers~\citep{DBLP:journals/corr/HazanKY17, sgdr_ilya_2016, cyclic_smith_2015, rg_wei_2018}, loss functions~\citep{focal_lin_2017,label_relax_yi_2018}, optimizers~\citep{liu2019variance, lookahead_zhang_2019} and
other network customizations~\citep{he_spatial_2015, hu2018gather, coordconv_liu_2018, zhang2017mixup, zhang2019making}. On the other hand, larger networks are being designed with increased depth and width to improve accuracy.

To further optimize these networks, there is an overemphasis on adding more complexity to either the network architecture or to the training procedure which can come at the cost of simplicity, explainability, inference time and energy efficiency. 
Moreover due to the energy inefficiency, the overall carbon footprint of the model increases significantly which has an adverse environmental impact~\citep{schwartz2019green}, for instance, training a common deep learning model can have a larger carbon footprint than half the life cycle of a car~\citep{strubell2019energy}.
In an effort to create a low carbon society to tackle climate change which is one of the major challenges faced by humanity today, researchers recently proposed using machine learning to decarbonize major pollution contributors such as the transport and the industrial sector~\citep{rolnick2019climate}. 
Paradoxically, this approach itself is becoming a major contributor to CO\textsubscript{2} emissions. Thus, addressing the energy efficiency and decarbonization of deep learning models during design and evaluation is vital. 

In conjunction with the aforementioned issues, the lack of a standardized methodology for experimentation has led to a non-rigorous evaluation of the utility of the different techniques.
The importance of this issue is also underscored by \textit{Bouthillier et al.},~\citep{unreproducible_bouthillier_2019}
\begin{center}
{\it``Reproducibility is not only about code sharing, but most importantly about experiment design''.}
\end{center}
Furthermore, the lack of a standardized pipeline resulting in unfair comparisons and the neglect of important variables such as energy consumption, design simplicity and evaluation of neural networks can play important roles in reinforcing "research bias", i.e. the process where scientists overlook or influence certain observations to report results in order to portray a certain outcome.

In this paper, we conduct an exhaustive and fair empirical study to evaluate architectural design and a wide spectrum of techniques that have been proposed to improve neural networks performance. Our findings echo the importance of the Occam's Razor principle and further demonstrate that following simple standardized practices can curtail the environmental impact with little to no drop in performance. We conclude that it is crucial to rethink the design and evaluation of neural networks to ameliorate the problem of research bias. Our main contributions are as follows:
\begin{itemize}
   \itemsep-.01em 
  \item We assess the performance of a diverse set of techniques on different datasets and tasks under common settings.
  \item We evaluate the effect of each technique on energy consumption and carbon emissions to estimate its environmental impact.
  \item We address the relationship between computational cost, energy efficiency, and performance gain.
  \item We highlight the necessity of a standard pipeline for neural network design and evaluation, and to curtail their carbon footprint.
  \item We show that following simple standardized practices can help in fair comparison of various techniques, thereby mitigating research bias.
\end{itemize}

\section{Related Work}
\label{sec:related_work}
As a consequence of the increasing need for computational resources for deep learning methods, some researchers are now interested in studying and documenting the effect of the carbon footprint of these techniques.
Strubell et al. ~\citep{strubell2019energy} study the energy consumption of different DL models during training.
They conclude that the energy required for training models has increased significantly in recent years to a point where the large carbon footprint is becoming a major concern, i.e. energy consumption of the average deep learning model training can produce more than half the carbon emissions of the entire life cycle of an automobile. Schwartz et al. ~\citep{schwartz2019green} argue that energy efficiency of DL models is as important as the accuracy and put forward multiple methods to estimate the carbon emissions of AI models. The authors contend that researchers should report energy consumption and floating point operations per model as an evaluation metric. Following this recommendation, we report these metrics in our study.

Inference, in general, contributes more towards the carbon footprint of a model during its life cycle compared to the training. To this end, a number of studies have proposed to optimize the inference process.
Wang et al. ~\citep{unified_wang_2019} provide a general framework to optimize inference on mainstream integrated GPUs.
Moons et al. ~\citep{moons2016energy} suggest the use of precision scaling for CNNs to reduce the overall energy consumption.
By representing weights and values that are least affected by quantization with a lower precision, they reduce the energy consumption. Dami et al. ~\citep{echoing_dami_2019} introduce the idea of "Data Echoing" whereby they optimize the pre-optimizer steps by reusing the idle upstream times of the GPUs to speed up training.
However, the focus of all the aforementioned methods is on hardware optimization and they do not investigate how the deep learning architectures themselves affect environment and performance related metrics.
After completion of this work, we became aware of the paper by Musgrave et al. ~\citep{musgrave2020metric} where they too emphasize fair comparisons and highlight flaws in experimental setups. However, the scope of their study is metric learning.
In this study, we focus on weighing the utility of different recent methods for classification and segmentation tasks on both the training and inference efficiency in terms of overall performance and the energy expended.

\section{Methodology}\label{sec:methodology}
Our methodology is designed to ensure that the DL algorithms are evaluated with minimal effect of the hardware and software used for training, testing and inference. First, we use the same computing environment for all our experiments (Intel 8700 and NVIDIA RTX 2080 Ti).
Second, we check all the techniques for each task (segmentation and classification) on the same network that performs competitively on well-known public datasets. We compare each technique with one another alongside the baseline network to determine the overall value these different techniques provide.
Third, we provide a comprehensive analysis by not only reporting performance, but also energy consumption and inference time.
For segmentation training, we report the mean intersection over union (mIoU) on training and validation sets and the overall energy consumed. For segmentation inference, the total number of parameters, floating point operations (FLOPs), inference time and energy per image are provided. 
For classification tasks, we report the same metrics as above with the exception of mIoUs, where classification accuracy is reported instead. In case of multiple possible hyperparameter settings, we report the best results for each technique.

\subsection{Base Network Selection}
\paragraph{Segmentation.}
BiSeNet \citep{yu_bisenet:_2018} (with a ResNet-18 backbone pretrained on ImageNet \citep{he2016deep}) is selected for the segmentation task. We use two datasets for segmentation training and inference, Cityscapes \citep{cordts_cityscapes_2016} and COCO-Stuff \citep{caesar2018coco}. For both datasets, we use a poly learning rate decay with a decay rate of 0.9, stochastic gradient descent (SGD) as optimizer and cross-entropy (CE) as the loss function. For Cityscapes, we train for 1000epochs with an initial lr of 0.025, batch size 16, and random rescaling between 0.5 to 2 with a base size of $1024$, followed by randomly cropped windows of size $512 \times 512$. For COCO-Stuff on the other hand, we train for 60 epochs with an initial lr of 0.01, batch size 12, and random rescaling between 0.5 to 2 with a base size of $640$, followed by randomly cropped windows of size $640 \times 640$.


\paragraph{Classification.}
ResNet-50 with no pre-training is chosen for the classification task. We train and test the network on CIFAR-100 \citep{alexcifar2012} and Tiny-ImageNet \citep{Pouransari2014TinyIV} datasets.
We use an initial $lr$ of 0.1 with a step-wise $lr$ decay, batch size 128, SGD optimizer and CE as the loss function with maximum number of epoch set to 200. 

These networks are selected as they are well-studied and known to perform well in their respective domains. 
Moreover, we are interested in real-time applications and want to reduce the time spent on multiple training runs.
The hyperparameter values for baseline experiments are the same as in the original publications.
Unless otherwise stated, we follow the standard training scheme given above. 

\subsection{Categories for Experimentation}
We divide the experiments into several subcategories. Grouping similar categories allows for a concrete comparison of different techniques based on the multiple reported metrics. A summary of our grouping is provided below:
\vspace{-0.5em}
\begin{multicols}{2}
\begin{itemize}
  \itemsep-0.2em 
  \item Architecture Modification
  \item Learning Rate Scheduler
  \item Data Augmentation
  \item Optimizer
  \item Loss Function
  \item Custom Nodes and Layer
\end{itemize}
\end{multicols}

\subsection{How do we measure energy consumption?}
We spawn a new thread running parallel to training or inference code. In this thread, we poll GPU power usage (using NVIDIA Management Library \citep{pynvml}) and integrate it over time, similar to the method described in \citep{pi2017gpu}. However, we only sample the power every 10ms as it was the shortest time giving consistent results. Since we use the same data pipeline and dataloader, the CPU power remains same throughout the experiments, making it essentially an offset that we exclude from our reports. To measure training energy, we run all experiments on the same machine for 10 epochs and scale it up to 1000 epochs. The only exception is the progressive resizing technique, where we run the measurement for all epochs as the number of computations varies during the run. 
To measure inference energy, we run 10000 forward passes with mini-batch size 1 of randomly generated "images". We repeat this procedure three times and report the average energy per image. Lacoste et al. ~\citep{lacoste2019quantifying} is used to calculate the $CO_2$ emissions in kg per week (kg/wk) with an RTX 2080 Ti. The $CO_2$ emissions are calculated for the models on 168 hours of video (1 week) recorded at $2048\times1024$ and 30 FPS.

\begin{table*}[tb]
\caption{Effect of different techniques on mIoU calculated for both training and validation sets, and total GPU energy used for training per run for segmentation on Cityscapes (with image size of $512\times512$) and COCO-Stuff (with image size of 640x640). All values that are same or better than the baseline are in bold and the best results are highlighted.}
\centering
\begin{tabular}{|l|ccc|ccc|}
\hline
 & \multicolumn{3}{c|}{Cityscapes} & \multicolumn{3}{c|}{COCO-Stuff} \\\cline{2-7}
Method & mIoU  & mIoU  & Energy & mIoU  & mIoU  & Energy \\
 & train. & valid. & train. & train. & valid. & train. \\
\hline\hline
Baseline & 84.4\% & 69.3\% & 19.6 MJ & 35.4\% & 27.1\% & 65.8 MJ\\
\hline
One-Branch & 83.5\% & 68.6\% & \bf 17.1 MJ & 34.48\% & 26.6\% & \bf {64.1 MJ} \\
\hline
Random Grad. & \bf \colorbox{lightergray}{85.1\%} & \bf \colorbox{lightergray}{69.7\%} & 19.7 MJ & \bf  {38.13\%} & 22.6\% &  68.9 MJ\\
Cyclic LR & 80.5\% & 67.6\% & \bf 19.5 MJ & 35.13\% & \textbf{27.2}\% &  \textbf{65.3 MJ}\\
Poly LR 1/2 Epochs & 82.3\% & \bf 69.3\% & \bf 9.8 MJ & 31.51\% & 26.2\% & \bf {35.6 MJ} \\
\hline
Prog. Resize & 80.3\% & 64.5\% & \bf \colorbox{lightergray}{8.0 MJ} & 29.08\% & 24.3\% & \bf \colorbox{lightergray} {30.3 MJ} \\
Mixup & 41.8\% & 68.1\% & 20.3 MJ & 16.19\% & 25.5\% & 84.5 MJ \\
\hline
RAdam & 82.5\% & 67.8\% & 20.1 MJ & 26.79\% & 21.9\% & 73.1 MJ \\
    LookAhead & \bf 84.9\% & 68.8\% & \bf 19.5 MJ & \bf \colorbox{lightergray} {39.82\%} & \bf {27.4\%} & 71.6 MJ \\
\hline
Label Relaxation & 81.6\% & 67.4\% & 33.7 MJ & 32.22\% & 24.9\% & 95.2 MJ \\
Dice Loss & \bf 85.6\% & 68.5\% & 25.7 MJ & \bf {36.67\%} & 26.7\% & 77.5 MJ \\
Focal Loss & 83.5\% & 68.5\% & 20.8 MJ & \bf {36.01\%} & \bf \colorbox{lightergray} {27.6\%} & 82.7 MJ \\
\hline
BlurPool & \bf 84.6\% & 67.3\% & 26.2 MJ & \bf{36.96\%} & \bf {27.4\%} & 91.1 MJ \\
SwitchNorm & 84.3\% & 68.7\% & 19.7 MJ & \bf{36.23\%} & \bf {27.3\%} & 76.8 MJ \\
Sp. Bottleneck & 79.8\% & 66.0\% & 20.8 MJ & 31.38\% & 24.9\% & 71.5 MJ \\
GE-$\theta$ & 84.3\% & 68.7\% & \bf 19.6 MJ & 32.15\% & 24.9\% & 72.2 MJ \\
GE-$\theta^{-}$ & 84.4\% & 68.3\% & \bf 18.8 MJ & 32.39\% & 25.2\% & 71.0 MJ \\
CoordConv & 83.1\% & 69.0\% & 24.3 MJ & 32.29\% & 25.3\% & 83.1 MJ \\
\hline
\end{tabular}
\label{tab:train_table_seg}
\end{table*}

\begin{table*}[tb]
\caption{Effect of different techniques on accuracy calculated for both training and validation/test sets, and total GPU energy used for training per run for classification on CIFAR-100 and Tiny-ImageNet. All values that are same or better than the baseline are in bold and the best results are highlighted.}
\centering
\begin{tabular}{|l|ccc|ccc|}
\hline
& \multicolumn{3}{c|}{CIFAR-100} & \multicolumn{3}{c|}{Tiny-ImageNet} \\\cline{2-7}
Method & Accuracy & Accuracy & Energy & Accuracy & Accuracy & Energy \\
& train. & test & train. & train. & valid. & train. \\
\hline\hline
Baseline & 99.96\% & 78.65\% & 2.40 MJ & 99.95\% & 87.15\% & 6.89 MJ \\
\hline
Random Grad. & 99.93\% & 78.07\% & 2.43 MJ & 98.76\% & 86.36\% & 8.16 MJ \\
Cyclic LR & 99.92\% & 74.08\% & \bf 2.35 MJ & 82.46\% & 74.65\% & 8.86 MJ \\
Poly LR 1/2 Epochs & 99.89\% & 77.26\% & \bf \colorbox{lightergray}{1.36 MJ} & 99.95\% & 86.49\% & \bf \colorbox{lightergray} {3.47 MJ} \\
\hline
Mixup & 99.35\% & \bf \colorbox{lightergray}{80.65\%} & 2.62 MJ & 96.6\% & \bf {87.27\%} & 7.81 MJ \\
\hline
RAdam & 99.50\% & 73.65\% & 4.31 MJ & \bf \colorbox{lightergray} {99.99\%} & 83.48\% & 8.91 MJ \\
LookAhead & \bf \colorbox{lightergray}{99.97\%} & \bf 79.10\% & 2.45 MJ & \bf \colorbox{lightergray} {99.99\%} & \bf \colorbox{lightergray} {88.03\%}& 9.51 MJ \\
\hline
Focal Loss & 99.60\% & 78.16\% & 2.41 MJ & 98.60\% & 86.55\% & 7.68 MJ \\
\hline
BlurPool & 99.93\% & \bf 79.44\% & 3.18 MJ & \bf {99.97\%} & 86.73\% & 12.11 MJ \\
SwitchNorm & 99.50\% & 77.12\% & 4.63 MJ & 99.92\% & 86.36\% & 17.14 MJ \\
Sp. Bottleneck & 99.93\% & 72.76\% & 4.25 MJ & 97.02\% & 85.22\% & 11.67 MJ \\
GE-$\theta$ & 99.94\% & \bf 78.98\% & 3.37 MJ & \bf{99.97\%} & \bf {87.43\%} & 22.24 MJ \\
GE-$\theta^{-}$ & 99.93\% & 78.57\% & 2.70 MJ & 99.95\% & 87.08\% & 7.01 MJ \\
CoordConv & 99.95\% & \bf 78.82\% & 2.79 MJ & 99.91\% & 86.61\% & 9.12 MJ \\
\hline
\end{tabular}
\label{tab:train_table_class}
\end{table*}

\section{Experimental Evaluation} \label{sec:evaluation}
We perform systematic experimentation to test each technique with all the major metrics in perspective. Each section is organized as follows: we first describe the relevant techniques, we then report a summary of the results. For techniques that do not modify the network architecture we do not report the inference metrics as they are same as the baseline.

\subsection{Architecture Modification}
\paragraph{Spatial Branch Ablation (One-Branch).}
Many recent CNNs use a multi-branch architecture for learning the spatial details and global context separately and then fusing these features to segment the images~\citep{poudel_contextnet:_2018, poudel_fast-scnn:_2019, yu_bisenet:_2018}. In such networks, the context branch is deep (including many layers), and usually performs calculations on an image that is scaled down by a certain factor either via strided convolutions or bilinear resizing, while the spatial branch consists of fewer layers to preserve the spatial detail. 
Yu et al. ~\citep{yu_bisenet:_2018} show that adding the spatial branch has a very small yet positive effect on mIoU and negligibly effects the inference time. However, no tests are conducted on how this branch can affect energy consumption. Here, we consider the effect of the spatial branch on the energy consumption.

Table \ref{tab:train_table_seg} shows that there is a negligible difference between the full model (baseline) and the One-Branch in terms of mIoU.
The validation mIoU is $0.7$ percentage points (pp) and $0.5$ pp below the baseline for Cityscapes (CS) and COCO-Stuff respectively, while both the required training energy and inference time (Table \ref{tab:infer_table_seg}) are significantly reduced. This highlights the importance of the trade-off between energy consumption and the performance, questioning the utility of such architectural complexity at the cost of energy and computational efficiency. 

\begin{table*}[t]
\caption{Per image inference metrics for segmentation (Cityscapes $512\times512$) and (COCO-Stuff $640 \times 640$). All values that are same or better than the baseline are in bold and the best results are highlighted. The $CO_2$ emissions are calculated for full resolution $2048\times1024$ images.}
\centering
\resizebox{\textwidth}{!}{
\begin{tabular}{|l|ccccc|ccccc|}
\hline
 & \multicolumn{5}{c|}{Cityscapes} & \multicolumn{5}{c|}{COCO-Stuff} \\ \cline{2-11}
 Method & Params & FLOPs & Time & Energy & $CO_2$ & Params & FLOPs & Time & Energy & $CO_2$\\ 
   & (M) & (G) & (ms) & (J) & (kg/wk) & (M) & (G) & (ms) & (J) & (kg/wk)\\ 
 \hline\hline
Baseline & 14.01 & 25.93 & 6.51 & 1.57 & 28.6 & 14.01 & 36.3 & 7.81 & 4.41 & 34.3 \\ \hline
One-Branch & \bf \colorbox{lightergray}{13.88} & \bf 22.85 & \bf \colorbox{lightergray}{5.07} & \bf \colorbox{lightergray}{1.26} & \bf \colorbox{lightergray}{22.2} & \bf \colorbox{lightergray}{13.88} & \bf {32.0 G} & \bf \colorbox{lightergray}{6.52} & \bf \colorbox{lightergray}{3.8} & \bf \colorbox{lightergray}{28.6} \\ \hline
BlurPool & \bf 14.01 & 34.16 & 11.7 & 2.892 & 51.4 & 14.01 & 47.8 & 13.9 & 6.6 & 61.1 \\
SwitchNorm & \bf 14.01 & 25.93 & 7.25 & 1.589 & 31.9 & 14.01 & 36.3 &  8.68 & 4.8 & 38.2 \\
Sp. Bottleneck & \bf 14.01 & \bf \colorbox{lightergray}{15.36} & \bf 6.51 & 1.608 & \bf 28.6 & 14.01 & \bf \colorbox{lightergray}{21.5} & \bf 7.76 & \bf {4.35} & \bf 34.1 \\
GE-$\theta$ & 21.88 & \bf 25.93 & 6.73 & 1.635 & 29.5 & 25.05 & 36.3 & 8.06 & \bf {4.40} & 35.4 \\
GE-$\theta^{-}$ & \bf 14.01 & \bf 25.93 & 6.71 & 1.711 & 29.5 & 14.01 & 36.3 & 8.04 & \bf {4.39} & 35.3 \\
CoordConv & 14.02 & 27.16 & 8.10 & 2.009 & 35.6 & 14.02 & 38.0 & 9.71 & 4.79 & 42.7 \\ \hline
\end{tabular}}
\label{tab:infer_table_seg}
\end{table*}

\begin{table*}[t]
\caption{Classification CIFAR-100 and Tiny-ImageNet. All values that are same or better than the baseline are in bold and the best results are highlighted. The $CO_2$ emissions are calculated for resolution $64\times64$ images.}
\centering
\resizebox{\textwidth}{!}{
\begin{tabular}{|l|ccccc|ccccc|}
\hline
 & \multicolumn{5}{c|}{CIFAR-100} & \multicolumn{5}{c|}{Tiny-ImageNet} \\ \cline{2-11}
 Method & Params & FLOPs & Time & Energy & $CO_2$ & Params & FLOPs & Time & Energy & $CO_2$\\ 
    & (M) & (G) & (ms) & (J) & (kg/wk) & (M) & (G) & (ms) & (J) & (kg/wk)\\ 
 \hline\hline
Baseline & 2.37 & 1.3 & \bf \colorbox{lightergray}{6.60} & 0.98 & \bf \colorbox{lightergray}{29.0} & 2.39 & 1.3 & \bf \colorbox{lightergray} {6.60} & \bf \colorbox{lightergray}{0.92} & \bf \colorbox{lightergray}{29.0} \\ \hline
BlurPool & \bf 2.37 & 1.93 & 6.70 & 1.0 & 29.4 & \bf 2.39 & 1.93 & 6.70 & 0.94 & \bf 29.4 \\
SwitchNorm & \bf 2.37 & \bf 1.3 & 23.1 & 2.4 & 101.5 & \bf 2.39 & \bf 1.3 & 22.9 & 2.2 & 100.7 \\
Sp. Bottleneck & \bf 2.37 & \bf \colorbox{lightergray}{0.86} & 7.32 & 1.1 & 32.1 & \bf 2.39 & \bf \colorbox{lightergray}{0.86} & 7.21 & 1.0 & 31.7 \\
GE-$\theta$ & 2.42 & \bf 1.3 & 6.97 & 1.0 & 30.6 & 2.59 & \bf 1.3 & 6.90 & 1.0 & 30.3 \\
GE-$\theta^{-}$ & \bf 2.37 & \bf 1.3 & 6.96 & 1.0 & 30.6 & \bf 2.39 & \bf 1.3 & 6.90 & 0.98 & 30.3 \\
CoordConv & 2.42 & \bf 1.3 & 6.66 & \bf \colorbox{lightergray}{0.97} & \bf 29.2 & 2.43 & \bf 1.3 & 6.70 & 0.94 & \bf 29.4 \\ \hline
\end{tabular}}
\label{tab:infer_table_class}
\end{table*}

\subsection{Learning Rate Scheduler}\label{sec:LR}
A number of learning rate manipulation techniques have been suggested over the years to improve training and inference for CNNs. Below we introduce and analyze the effect of a number of learning rate optimization techniques.

\paragraph{Random Gradient.}
Random gradient~\citep{rg_wei_2018} multiplies the learning rate for each mini-batch by a random number sampled from $U([0,1])$. The authors claim this minimizes fluctuations in the optimization process.
This technique has been shown to perform well in many fields.

\paragraph{Cyclic Learning Rate.}
Cyclic learning rate (CLR) \citep{cyclic_smith_2015}, developed to remove the need for a learning rate hyperparameter, is based on cyclically changing the learning rate throughout the training. The variant giving the best result is named \emph{triangular2}, which varies the learning rate linearly and halves the upper limit every cycle. The authors show that it can improve the training and overall accuracy of different types of neural networks. For segmentation, we tried multiple hyperparameter settings and report the best results achieved with \texttt{base\_lr}=0.0001 and \texttt{base\_lr}=0.00004, \texttt{max\_lr}=0.01 and \texttt{max\_lr}=0.004, and \texttt{stepsize}=50 and \texttt{stepsize}=3 for Cityscapes and COCO-Stuff respectively. For classification, we set a \texttt{base\_lr}=0.0001, \texttt{max\_lr}=0.01 and a \texttt{stepsize}=20 epochs for both datasets.

\paragraph{Poly LR with \textonehalf~Maximum Epochs.}
In our base experiment, learning rate follows a polynomial curve. This means that the rate at which the learning rate decreases with each epoch is inversely proportional to maximum number of epochs. We test how having a smaller number of maximum epochs affects training and validation. To ensure that this result is not due to the variability in experiments we run three experiments and report the average mIoU.

\paragraph{Results.}
In Table~\ref{tab:train_table_seg}, we show that for segmentation compared to the baseline, random gradient experiments yield an improvement in mIoU for CS, but a decrease in COCO-Stuff mIoU at the cost of a small increase in energy. On the other hand, for cyclic learning rate, a significant drop in mIoU is observed for CS, whereas a minor increase in mIoU is observed for COCO-Stuff. Running the network for half the number of epochs with a poly $lr$ reduces the energy consumption (19.6 MJ to 9.8 MJ for CS and 65.8 MJ to 35.6 MJ for COCO) with little to no drop in validation mIoU. This emphasizes that proper experimental design can play an important role in environmental impact.

For classification, we observe that neither CLR nor random gradient affect the energy consumption significantly for CIFAR-100 while for the larger training run with Tiny-ImageNet the training energies go up from the baseline. The overall test/validation accuracy with random gradient is slightly lower, while CLR reports considerably lower accuracy than the baseline. For half the number of maximum epochs with poly $lr$, we observe that the energy consumption is almost halved while the test/validation accuracy is $1.39$ pp and $0.66$ pp lower than the baseline for CIFAR-100 and Tiny-ImageNet respectively. However, we noted that our evaluation matches that of the original CLR publication, the only difference being that they use a constant $lr$ for the base experiment.

\begin{figure*}[t]
\centering
\begin{tabular}{cc}
\subfigure{\includegraphics[width=0.49\linewidth]{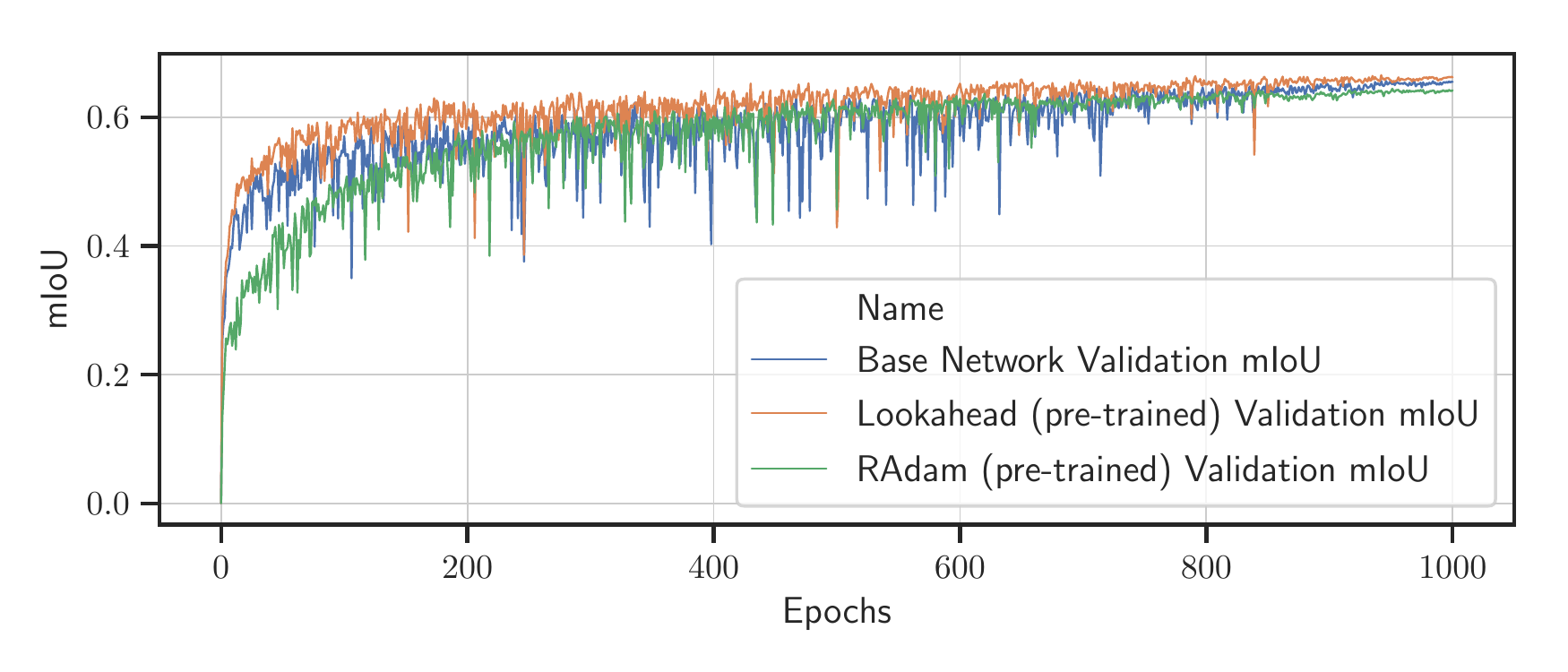}} & \subfigure{\includegraphics[width=0.49\linewidth]{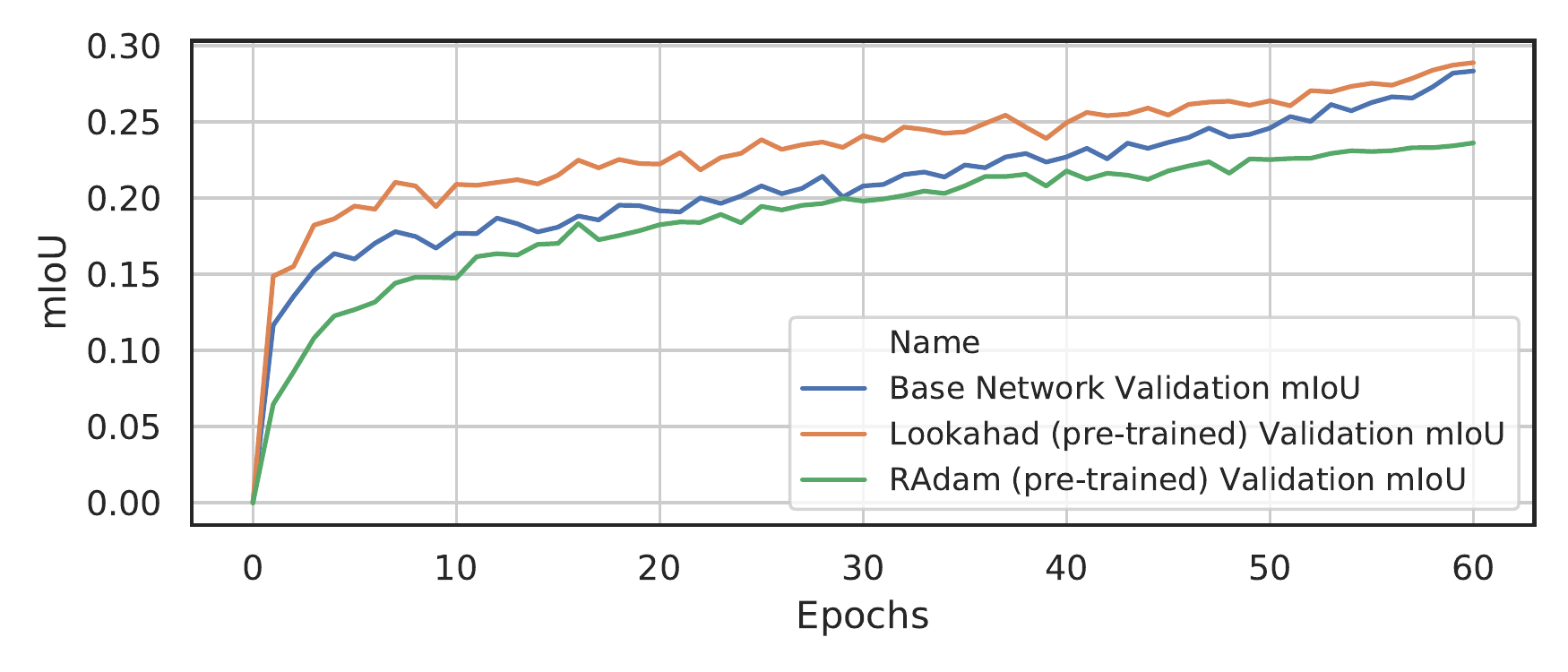}} \\
\subfigure{\includegraphics[width=0.49\linewidth]{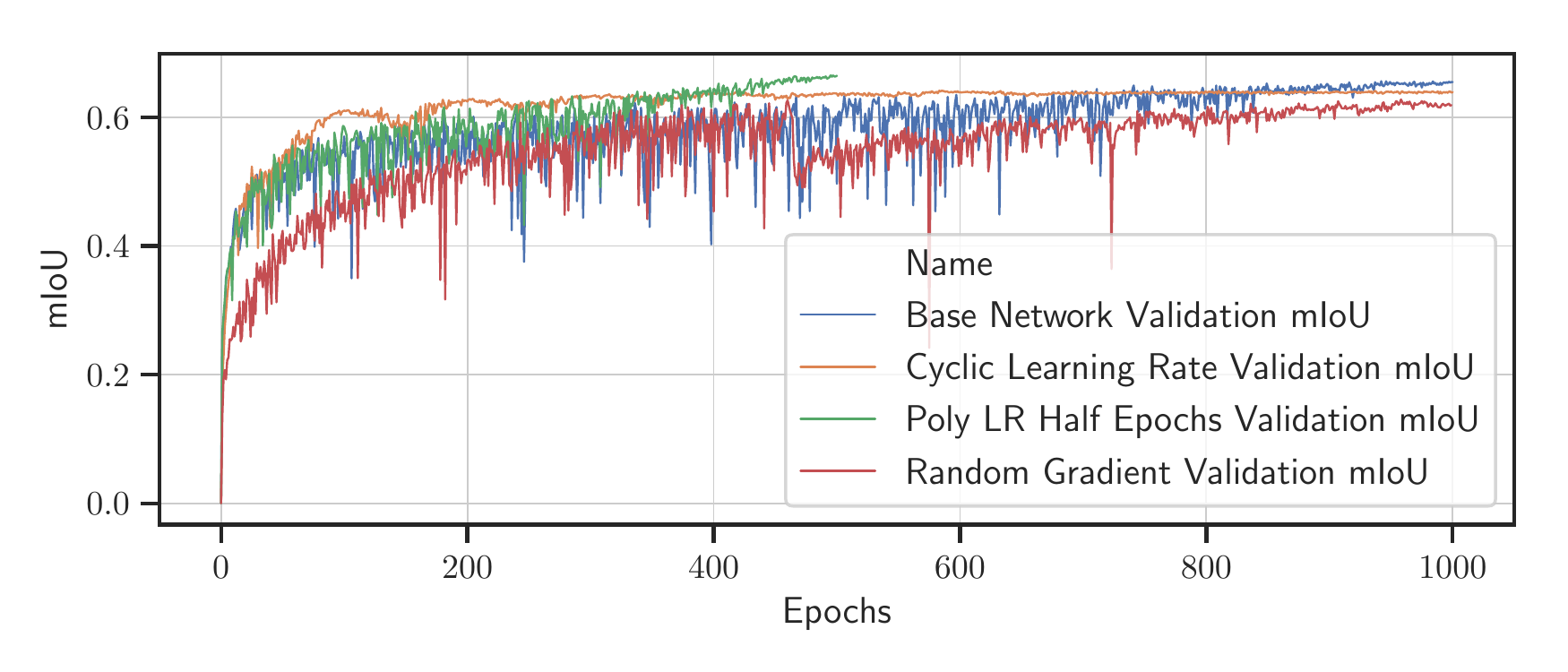}} & \subfigure{\includegraphics[width=0.49\linewidth]{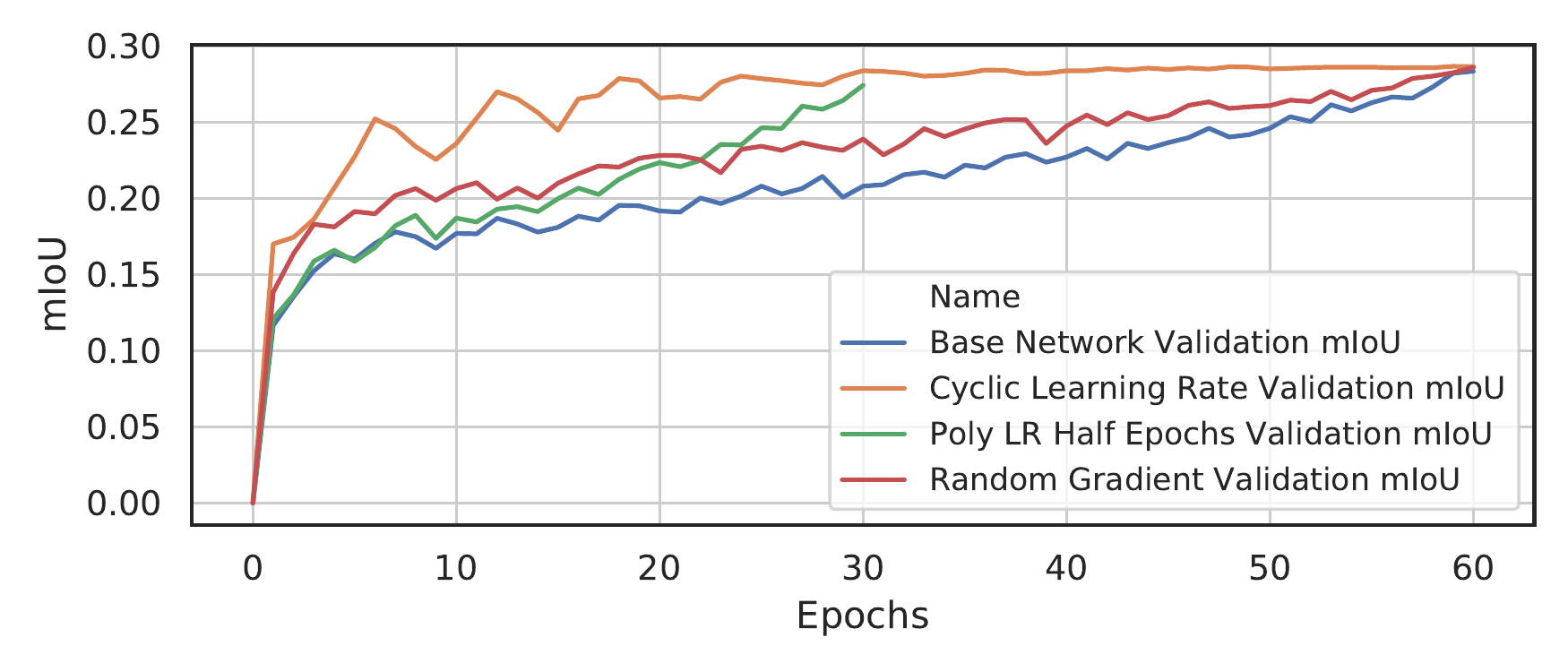}} \\
\end{tabular}
\caption{\textbf{(Top)} The validation mIoU (calculated on square center crops) profiles for different optimizers(Cityscapes on \textbf{Left} and COCO-Stuff on \textbf{Right}) indicate that LookAhead optimizer has a better convergence than the baseline while the convergence for RAdam is slower than the baseline. \textbf{(Bottom)} The validation mIoU for different learning rate modifications(Cityscapes on \textbf{Left} and COCO-Stuff on \textbf{Right}). Poly LR with half epochs still gives competent results, even outperforming the remaining methods on Cityscapes. Cycles are visible for CLR. Values don't exactly match the Table as mIoU was calculated only on the square center crops.}
\label{fig:lr_val}
\end{figure*}

\subsection{Data Augmentation}
Different types of data augmentation techniques have been applied to neural networks to deal with the scarcity of training data. These techniques reuse data after applying image processing techniques, such as random crops, random flipping, resizing, brightness and contrast modification. They have, in several cases, shown to improve the robustness of CNNs~\citep{cubuk2018autoaugment}. We run our baseline experiment with the aforementioned techniques.

\paragraph{Progressive Resizing.}
Progressive resizing is a technique where the training is sped up by training on downsized images for a given number of epochs, followed by training progressively on relatively larger images until the point where final epochs are done on images of original size~\citep{arani2019rgpnet, howard2018fastai}. This has been shown to reduce training times significantly.
We run our experiment with image size $\frac{1}{8}$ of the original image, and upscale the image by a factor of 2 after every 250 epochs for Cityscapes, and 15 epochs for COCO-Stuff. This is done until we get the original image size, which is kept constant for the rest of the training.

\paragraph{Mixup.}
Mixup~\citep{zhang2017mixup} is a simple data augmentation scheme, which trains the network on convex combinations of pairs of data points and the corresponding one-hot representations of their labels, to improve generalization of deep neural networks. 
\begin{equation}
  \label{eqn_1}
  \begin{aligned}
    \tilde{x} &= \lambda x_i + (1-\lambda) x_j\\
    \tilde{y} &= \lambda y_i + (1-\lambda) y_j
  \end{aligned}
\end{equation}
Where $0\leq\lambda\leq1$, $x_i, x_j$ are raw input vectors, and $y_i, y_j$ are one-hot label encodings.
Here, we evaluate how the mixup affects the overall training process.

\paragraph{Results.}
For segmentation, progressive resizing results in an inferior mIoU but considerably better training energy (see Table~\ref{tab:train_table_seg}), while for mixup the negative effect on mIoU is less pronounced with a slight increase in energy consumption for both datasets.
For classification, we observe that mixup leads to a significant improvement in test/validation accuracy with almost identical energy consumption for CIFAR-100 and increased consumption for Tiny-ImageNet. We do not employ progressive resizing for classification as the CIFAR-100 and Tiny-ImageNet images are too small ($32\times 32$) to retain enough information after resizing.

\subsection{Optimizer}
DNNs are trained using variants of stochastic gradient descent. Here, we evaluate recently proposed optimizers which claim to improve the performance of SGD.

\paragraph{RAdam Optimizer.}
The RAdam optimizer~\citep{liu2019variance} employs an adaptive learning rate which is rectified to ensure a consistent variance. It effectively provides an automated warm-up custom tailored to the current dataset.
Warm-up learning rate strategy sets up a small learning rate in the starting phase of training to cater for the undesirably large variance. After a specific number of epochs, the learning rate is stepped up to a larger value.
\begin{equation}
lr = \left\{
\begin{array}{ll}
      x_e \cdot lr_0 & e < n \\
      x_n \cdot lr_0 & e\geq n \\
\end{array}
\right. \
\end{equation}
where $n$ is the number of threshold epochs for the warm-up, $x_i$ is the learning rate modifier for $i$th epoch, $e$ is the current epoch, and $lr_0$ denotes the base learning rate.
In case of the spiking learning rate warm-up we have,
\begin{equation}
lr = \left\{
\begin{array}{ll}
      x \cdot lr_0 & e~mod~n = 0 \\
      lr_0 & else \\
\end{array}
\right. \  
\end{equation}
RAdam was run with default hyperparameters used by Liu et al. \citep{liu2019variance} while keeping all other settings of the optimizer same as the baseline.
\paragraph{LookAhead Optimizer.}
LookAhead optimizer uses "fast weights" multiple times using another standard optimizer before updating the "slow weights" in the direction of the final fast weights~\citep{lookahead_zhang_2019} and has been shown to reduce the variance, resulting in better convergence and versatility with little hyperparameter tuning.
We run LookAhead with outer optimizer looking ahead every 5 iterations and interpolation coefficient $\alpha = 0.5$. Inner optimizer has the same optimization routine and learning rate schedule as the baseline. Same hyperparameters for LookAhead were chosen as those by Zhang et al. \citep{lookahead_zhang_2019} for ImageNet data. 
\paragraph{Results.}
For segmentation, Table \ref{tab:train_table_seg} shows that RAdam produces an mIoU that is $1.5$ pp and $5.1$ pp lower than the baseline on Cityscapes and COCO-Stuff respectively. On the other hand, it consumes around $0.5$ MJ more energy on Cityscapes, and $7.3$ MJ more energy on COCO-Stuff. LookAhead optimizer performs relatively better with the reported mIoU only $0.5$ pp below the Cityscapes baseline with similar energy consumption, and $0.3$ pp above the COCO-Stuff baseline with $5.8$ MJ more energy consumption. Figure \ref{fig:lr_val} shows that lookahead optimizer improves the convergence over the baseline (SGD).
For classification, in Table \ref{tab:train_table_class} we observe a test accuracy drop of $5.0$ pp for RAdam and an increase of $0.45$ pp for lookahead on CIFAR-100, and a validation accuracy drop of $3.67$ pp for RAdam and an increase of $0.88$ pp for lookahead on Tiny-ImageNet, indicating that the former reduces accuracy while the latter improves it. Lookahead has almost the same energy as the CIFAR-100 baseline but consumes more energy on Tiny-ImageNet, while the RAdam optimizer consumes significantly more energy on both datasets.

\subsection{Loss Function}
There is no loss function that suits all deep learning algorithms. The choice of loss function can have a significant effect on the overall performance of a deep learning model as well as the speed of convergence. 

\paragraph{Boundary Label Relaxation.}
The boundary label relaxation \citep{label_relax_yi_2018} leverages the idea that segmentation boundaries close to one another can easily be classified as one or the other without any marked impact on the quality of segmentation. Based on this, boundary label relaxation calculates a soft cross-entropy loss value, where the summation of the probabilities of all the neighbouring classes at a given pixel is used as the final class probability for that pixel.
\begin{equation}\label{eqn_2}
    P(C) = P(A\cup B) = P(A) + P(B)
\end{equation}
\begin{equation}\label{eqn_3}
    L_{Relax} = -log\sum_{C \in N}{P(C)}
\end{equation}
where A and B are the mutually exclusive classes and $P(.)$ represents the softmax class probability.
We perform the experiment with label relaxation where a $3\times3$ kernel is used to check for neighbouring pixels. 

\paragraph{Soft Dice Loss.}
The dice coefficient gives a measure of overlap between two sample sets. Dice loss applies the dice coefficient to measure the overlap between the softmax output and the ground-truth \citep{milletari2016v, Sudre_2017}.
\begin{equation}\label{eqn_4}
    L_{Dice}(p,\hat{p}) = 1-\frac{2\sum_{pixels}p\hat{p}}{\sum_{pixels}p+\sum_{pixels}\hat{p}}
\end{equation}

We use,
\begin{equation}\label{eqn_5}
    L_{SoftDice} = L_{CE} + \gamma L_{Dice}
\end{equation}
where 
$\gamma$ is a scaling factor for the dice loss. When the network is trained with dice loss alone, significantly lower performance is observed. Therefore, we tested with multiple values of $\gamma$ and got the best performance with $\gamma=1$.

\paragraph{Focal Loss.}
Lin \textit{et al.} \citep{lin2017focal} introduced focal loss after observing that two stage detectors are more accurate because of the inherent extreme foreground-background class imbalance. They introduce focal loss to address this class imbalance by focusing more on hard negatives and down-weighing the easier samples. In this study, we adapt the focal loss for segmentation by treating each pixel as a separate sample.
\begin{equation}\label{eqn_6}
    L_{Focal}(p,\hat{p})=-\sum_{pixels} [ ((1-\hat{p})^\gamma \log(\hat{p})) \cdot p ],
\end{equation}
where $\hat{p}$ is softmax output vector, $\cdot$ is a dot product, and p is a one-hot label vector. We train with $\gamma=2$ as according to author it yields the best results.

\paragraph{Results.}
For segmentation, the boundary label relaxation lowers the mIoU by $1.9$ pp and $2.2$ pp for CS and COCO-Stuff respectively while consuming significantly more energy. For dice loss, we get a lower mIoU ($-0.8$ pp for CS and $-0.4$ pp for COCO-Stuff) relative to the baseline and $6.1$ MJ and $11.7$ MJ increase in energy consumption. Additionally, an improvement (~$1.2$ pp for both datasets) in training mIoU is observed. For the focal loss, we observe a drop of $0.8$ pp and an increase of $0.5$ pp in validation mIoU with more energy consumption for CS and COCO-Stuff respectively.
For classification, we test only focal loss and observe that it marginally reduces the test/validation accuracy ($0.49$ pp and $0.60$ pp for CIFAR-100 and Tiny-ImageNet respectively).

\subsection{Custom Nodes and Layers}
A number of studies have been carried out to improve the neural networks performance by introducing new custom nodes and layers.

\paragraph{BlurPool (Shift-Invariant Pooling).}
Zhang~\citep{zhang2019making} introduced a new way to perform the pooling operation for resizing features, called BlurPool. This challenges the common assumption that CNNs are inherently shift-invariant. This is evident from the fact that even small translational perturbations (e.g. due to max pooling, resizing etc.) of an image can produce a drastically erroneous result. To counter the effect of this uncertainty, anti-aliasing is used in conventional image processing. The study uses the same idea for neural networks by adding anti-aliasing filters prior to downsampling. We run the BlurPool experiment by modifying all downsampling operations with BlurPool layers of kernel size $5\times5$ while keeping the strides same as the original downsampling operation. 

\paragraph{Switchable Normalization.}
A number of normalization techniques have been put forward in recent years, such as batch normalization (BN) \citep{ioffe2015batch}, instance normalization (IN) \citep{ulyanov2016instance} and layer normalization (LN) \citep{ba2016layer}. The relative performance of each technique depends on the task at hand. Switchable Normalization (SN) \citep{luo2018differentiable} learns to combine different normalization techniques. This experiment is carried out with all the batch normalization layers replaced by switchable normalization layers. For segmentation, we use switchable normalization everywhere except the backbone. We report these results in Table \ref{tab:train_table_seg} and \ref{tab:infer_table_seg}.

\paragraph{Spatial Bottleneck.}
Spatial bottlenecks \citep{peng2018accelerating} decomposes convolutions into two stages for computational efficiency. The first stage reduces the spatial computation  using stride $k$ for the convolution layer. The second stage uses transposed convolution layer with the same stride $k$ to recover the original size. According to authors, this reduces computations by a factor of $2/k^2$, and can hence be used to reduce computation with a tolerable reduction in accuracy. The spatial bottleneck experiments are run with the ResNet18 backbone modified to include the ResBlocks with spatial bottleneck modules instead of the standard convolution and deconvolution. 

\paragraph{Gather-Excite Operators.}
Hu et al. \citep{hu2018gather} introduced gather-excite (GE) operators to better capture long range dependencies and interactions across an image. The gather operator aggregates the features with a large receptive field and then the excite operator redistributes the condensed information to the local features without adding much to the computational overhead. We refer to the GE modules with trainable parameters as GE-$\theta$ and the non-trainable modules as GE-$\theta^{-}$. GE-$\theta$ and GE-$\theta^{-}$ experiments are also carried out by attaching these modules to the outputs of each of the ResBlocks in the ResNet18 backbone.

\paragraph{Coordinate Convolution.}
The Coordinate Convolution (CoordConv), introduced by Liu et al. \citep{coordconv_liu_2018}, concatenates pixel coordinates to the input tensor before a convolution, to propagate spatial information to deeper layers. The coordinates are added as additional channels having x, y and radial coordinates correspondingly. The authors claim that the technique improves localization accuracy for GANs and classification tasks. For the CoordConv experiment, we concatenate the Cartesian and radial features just to the first layer of the backbone and the first layer of the spatial path. We use the same range of co-ordinates for training and testing, despite cropping half the pixels in training as this yields the higher mIoU. 


\paragraph{Results.}

In segmentation training for both Cityscapes and COCO-Stuff, we observe similar trends with a few exceptions and notable results.
For Cityscapes, we observe that Random Gradient yields the best training and validation mIoU, that is $0.7$ pp and $0.4$ pp higher, respectively, than the baseline. This improvement could be attributed to the lower lr for training as Random Gradient multiplies the lr with $ 0 < \alpha < 1$. Another result of note is that for $1/2$ the number of epochs, we get the same validation mIoU, indicating that training for more epochs (in the baseline) leads to overfitting. For COCO-Stuff on the other hand, we observe that focal loss yields the best result as the data balancing from focal loss plays a greater role for the larger and more imbalanced dataset. The most notable results for training energy we observe are on progressive resizing and $1/2$ epoch run for both the datasets. Even though progressive resizing runs for twice the number of epochs, it consumes less energy than the $1/2$ epoch run, indicating that image size has a greater impact on training energy than the number of epochs.

For both the classification datasets, most of the train accuracy values are over $99\%$ with much lower test accuracy, indicating that all the methods overfit on the training set. For test accuracy, we observe that methods such as Mixup and Lookahead yield the best results. Lookahead optimizer performs well as it reduces the variance 
of the gradients of randomly picked batch sizes, while Mixup improves performance because of the strong regularization it provides. In terms of training energy, as expected, running half the number of epochs leads to lowest consumption.

For segmentation inference, 
an interesting observation we make is that the inference speed is not directly proportional to FLOPs as variables such as data parallelism and memory transfer can play a role in overall inference speed. Additionally, note that the difference in inference times observed for COCO-Stuff and Cityscapes is because we report the results for them at different resolutions ($640\times640$ and $512\times512$ respectively).
In case of classification inference, we observe that all the methods are slower than the baseline and thus possess a larger carbon footprint than the baseline.

\begin{figure*}[h]
\centering
\begin{tabular}{cc}
\subfigure[]{\includegraphics[width=0.48\linewidth]{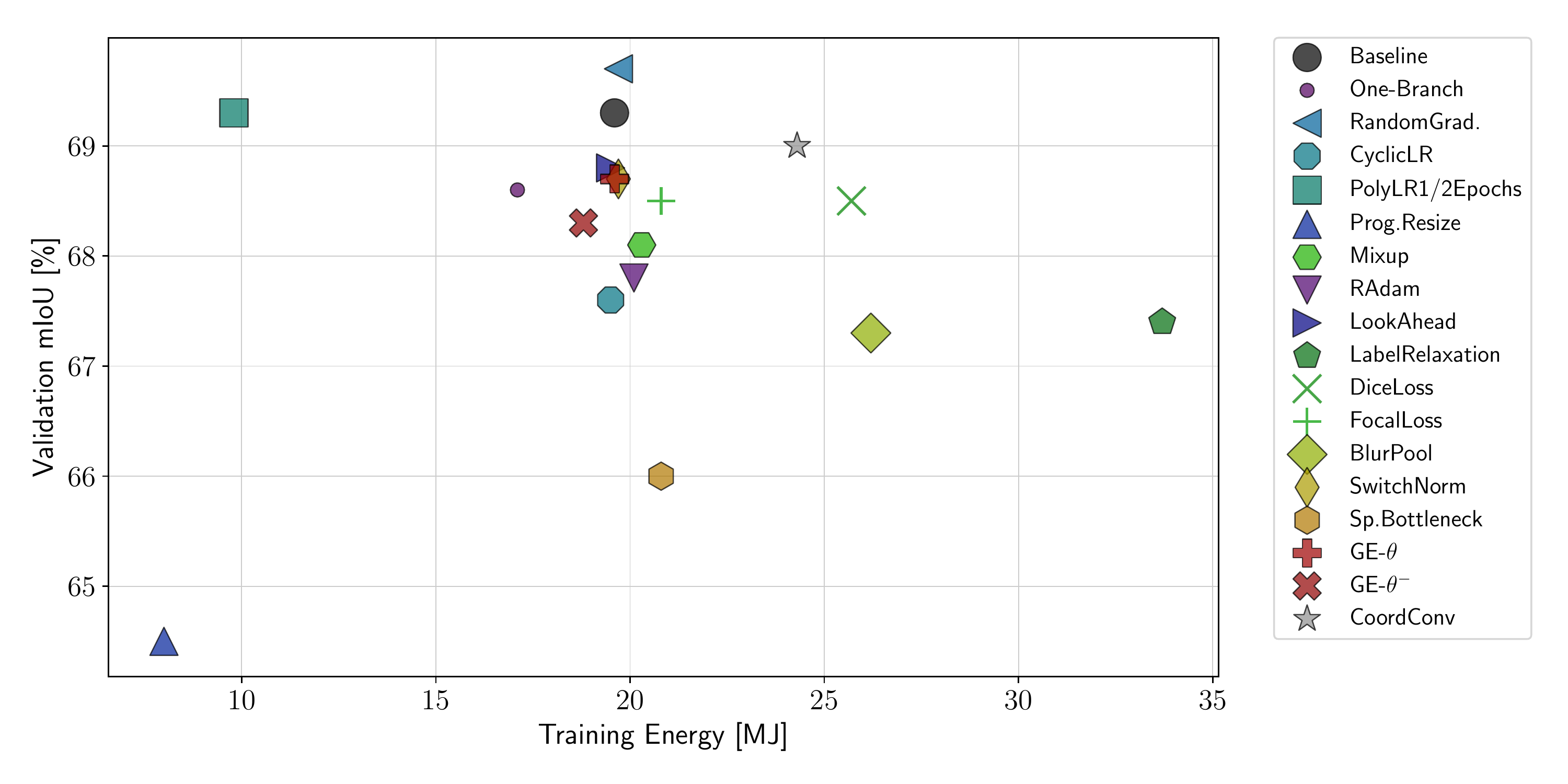}} &
\subfigure[]{\includegraphics[width=0.45\linewidth, height=0.24\linewidth]{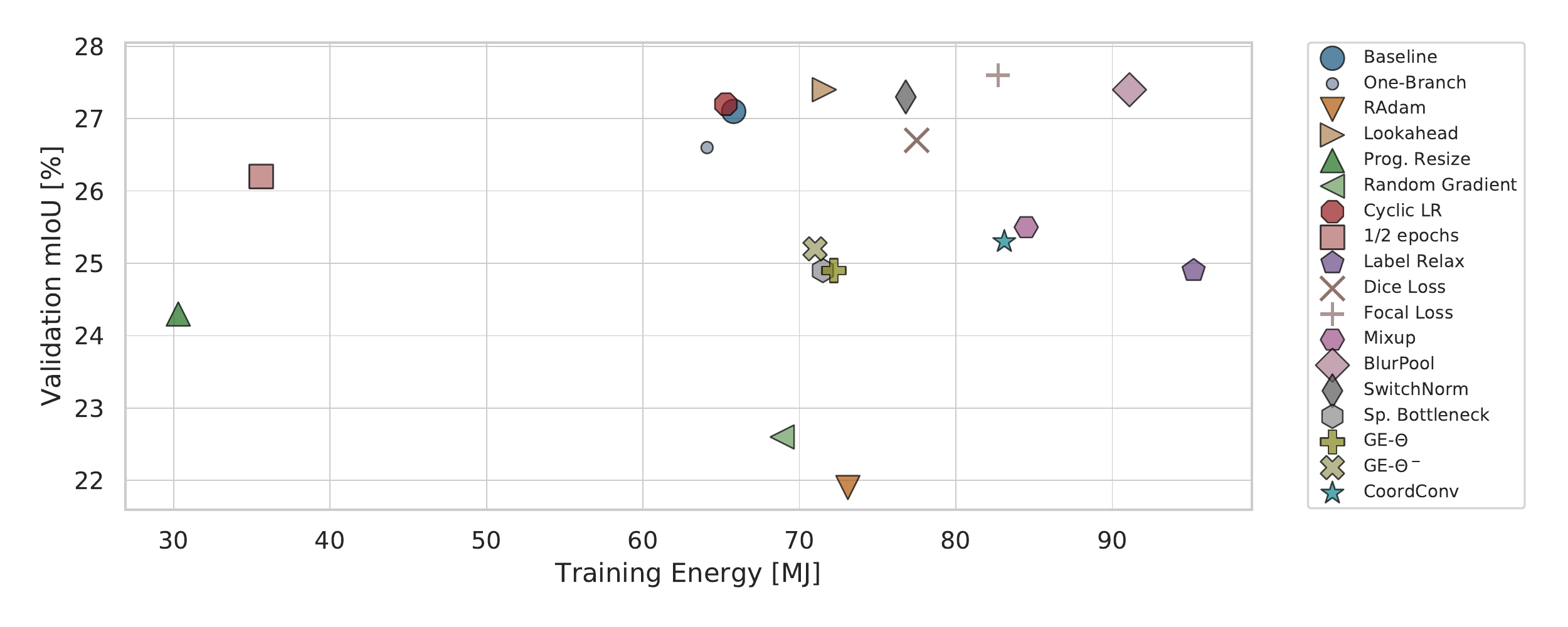}}
\end{tabular}
\caption{Validation mIoU vs. training energy for each technique on \textbf{(a)} Cityscapes and \textbf{(b)} COCO-Stuff: no correlation between the training energy and validation mIoU is observed.}
\label{fig:plot_validation_miou_vs_training_energy}
\end{figure*}

\begin{figure*}[h]
\centering
\subfigure[]{\includegraphics[width=0.7\linewidth]{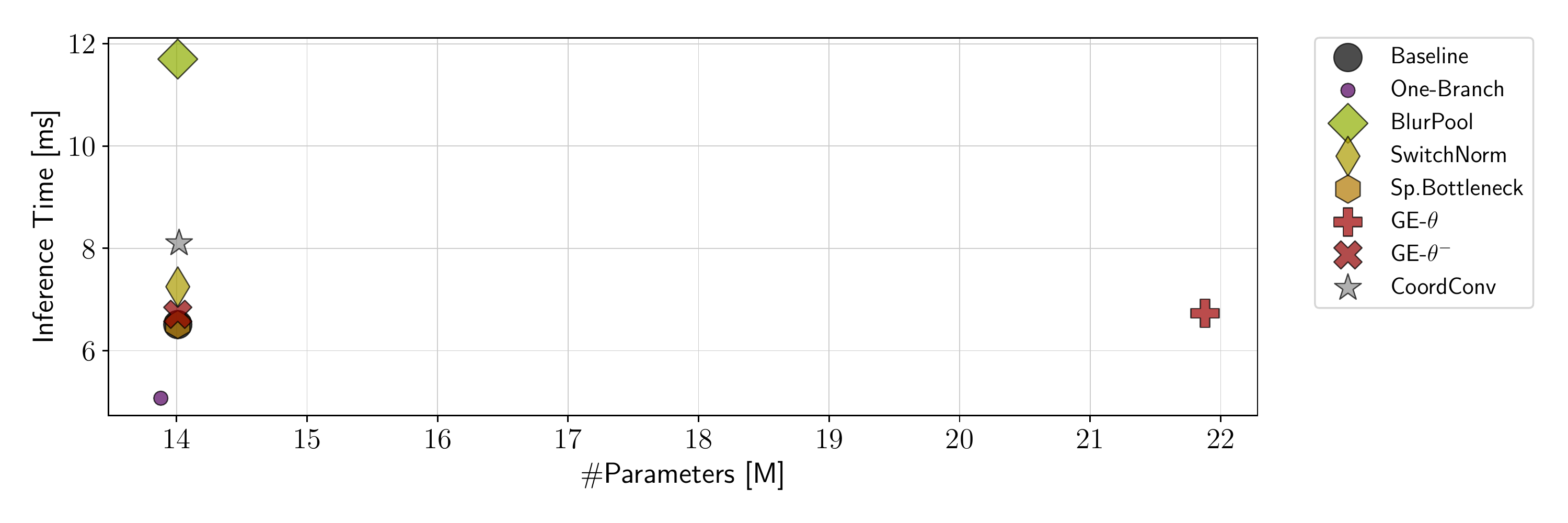}} \\
\subfigure[]{\includegraphics[width=0.7\linewidth]{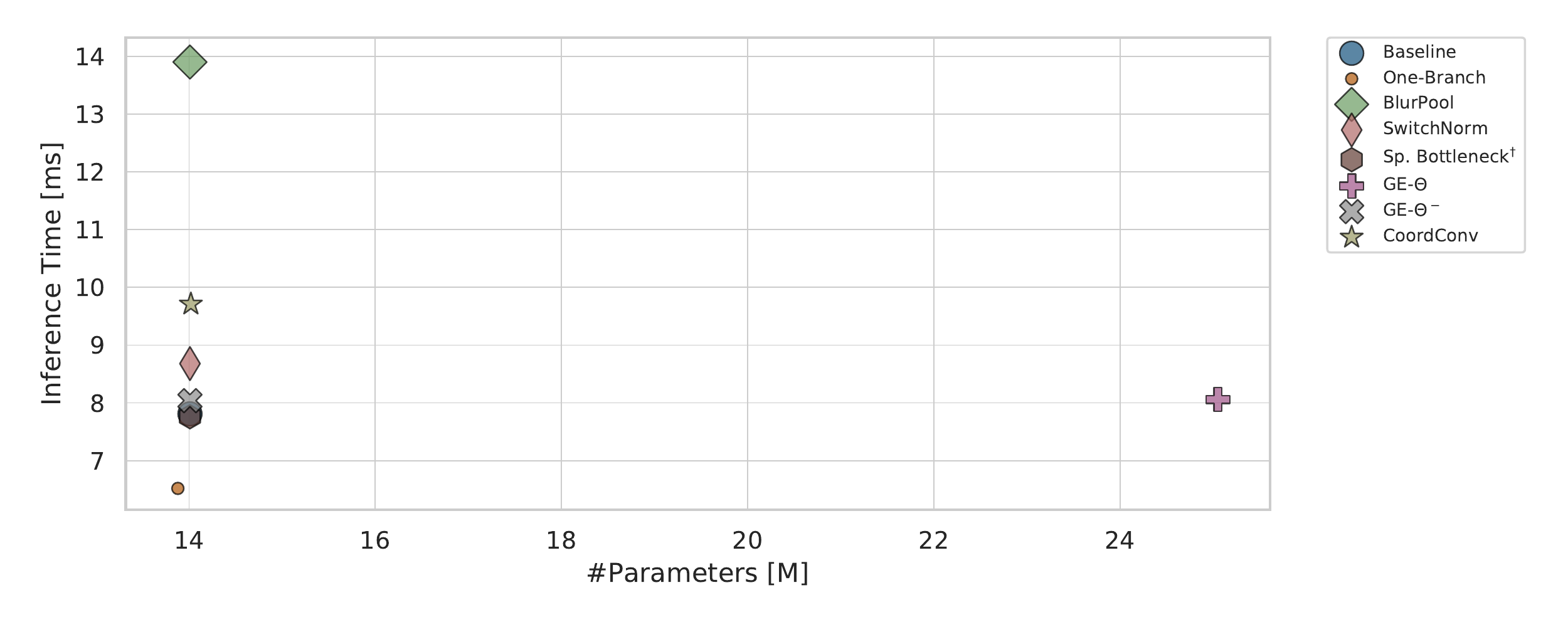}}
\caption{Inference Time vs. Number of parameters for each technique on \textbf{(a)} Cityscapes and \textbf{(b)} COCO-Stuff: The number of parameters is not the only determinant for inference time. This highlights the need to make effective use of parameters in NN design, for instance, ensuring minimal memory overhead, and high parallelization.}
\label{fig:plot_inference_time_vs_parameters}
\end{figure*}


\section{Discussion}\label{sec:Discussion}
This study empirically evaluated architectural design and a broad spectrum of methods proposed to improve CNNs performance to highlight the issue of research bias in deep learning.
Training an average deep learning model can have an immense impact on the environment. For instance, a model can have as much as more than half the carbon footprint of a car's entire life-cycle \citep{strubell2019energy}. On the other hand, researchers have recently proposed the use of machine learning models to tackle climate change \citep{rolnick2019climate}. Paradoxically, these models themselves can have a considerable environmental impact. We therefore perform an extensive evaluation in order to assess the suitability of different techniques for distinct tasks.
We used BiSeNet(ResNet-18 backbone) as our baseline for segmentation task, and observed that removing the spatial branch had a modest impact on the accuracy but considerably improved the carbon footprint and inference time. This again emphasizes the importance of model simplicity in addition to the overall performance during the design process. On the other hand, for training, we also showed that by simply reducing the maximum number of epochs, we can get the same accuracy at half the carbon cost. We evaluated a wide variety of other techniques, such as CoordConv, anti-aliasing for shift-invariance, optimizers such as LookAhead, RAdam, and loss functions such as Focal Loss, Dice Loss, and observe that a majority of these techniques have a negligible effect on the overall network performance. Moreover, some of these techniques considerably increased the complexity, carbon footprint and inference times. From the results in Figure \ref{fig:plot_validation_miou_vs_training_energy}, we could not observe a clear trade-off between accuracy and energy savings. In addition, Figure \ref{fig:plot_inference_time_vs_parameters} illustrated that the number of parameters is not the only determinant of the inference time.
These results highlighted the need to design networks with a broader spectrum of variables under consideration. More extensive demonstrations of the lack of correlation between these variables can be found in Appendix \ref{Appendix}.
While machine learning competitions such as Kaggle have been a great boon towards advancing research, they have exacerbated the issue of research bias by causing researchers to overlook several important variables such as environmental impact, and versatility. This appears to be a good example of Goodhart's Law which states that "When a measure becomes a target it ceases to be a good one" \citep{strathern_1997}. While we report our results on vision based applications, the same conclusions may extend to other DL intensive applications such as NLP \citep{bender2021dangers, ethayarajh2020utility}.
In this study, we call the community's attention to these important variables, and ask them to rethink the experimental design for deep learning; first by taking these important variables into account, and second by following a standard pipeline. Simple standardized practices can lead to significant reduction in environmental impact, as evidenced by Section \ref{sec:LR}. We also suggest the research community to take a step back and focus on the bigger picture when designing the networks, instead of targeting minute improvements with narrow applicability at the cost of simplicity, versatility, and energy.

\bibliographystyle{unsrtnat}
\bibliography{references}

\clearpage
\appendix
\section{Appendix}\label{Appendix}
Figures provided below further demonstrate the lack of correlation between accuracy/mIoU, and training energy and $CO_2$ emissions. 
\FloatBarrier
\begin{figure*}[!h]
\centering
\subfigure(a){\includegraphics[width=0.7\linewidth]{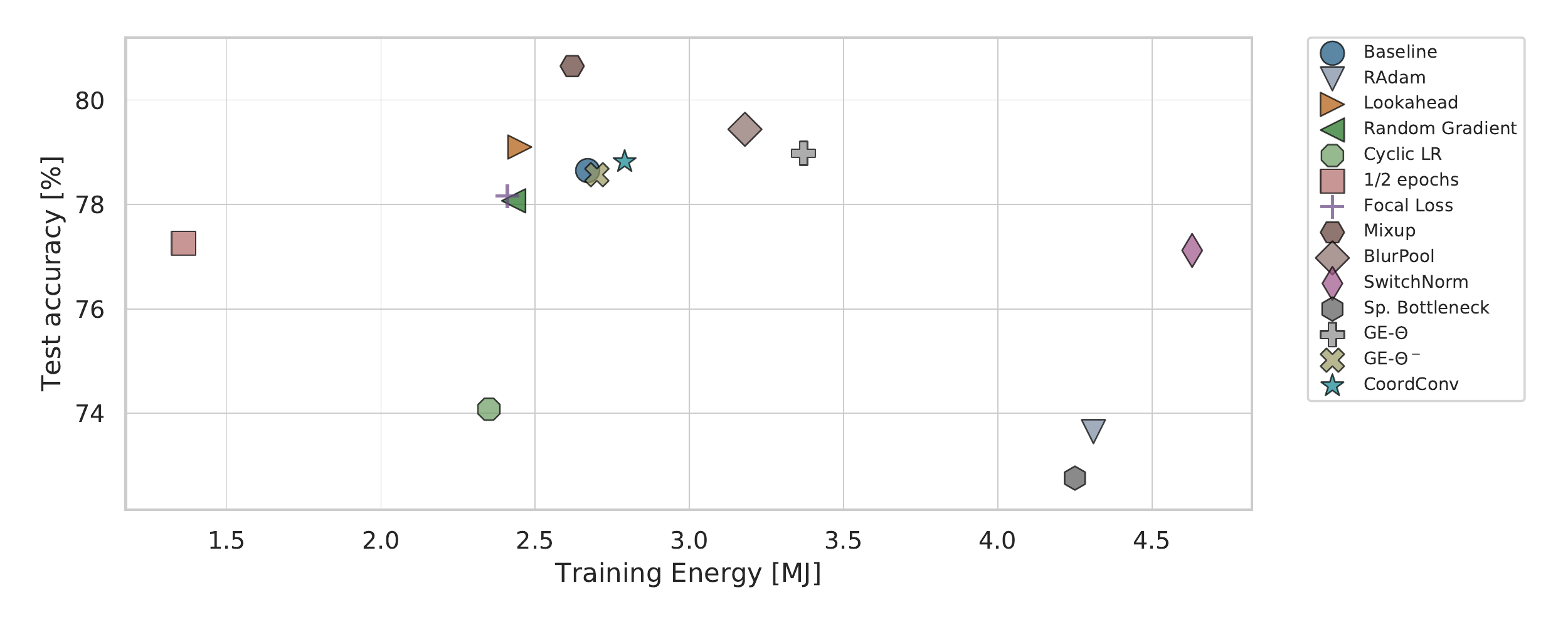}} \\
\subfigure(b)\includegraphics[width=0.7\linewidth]{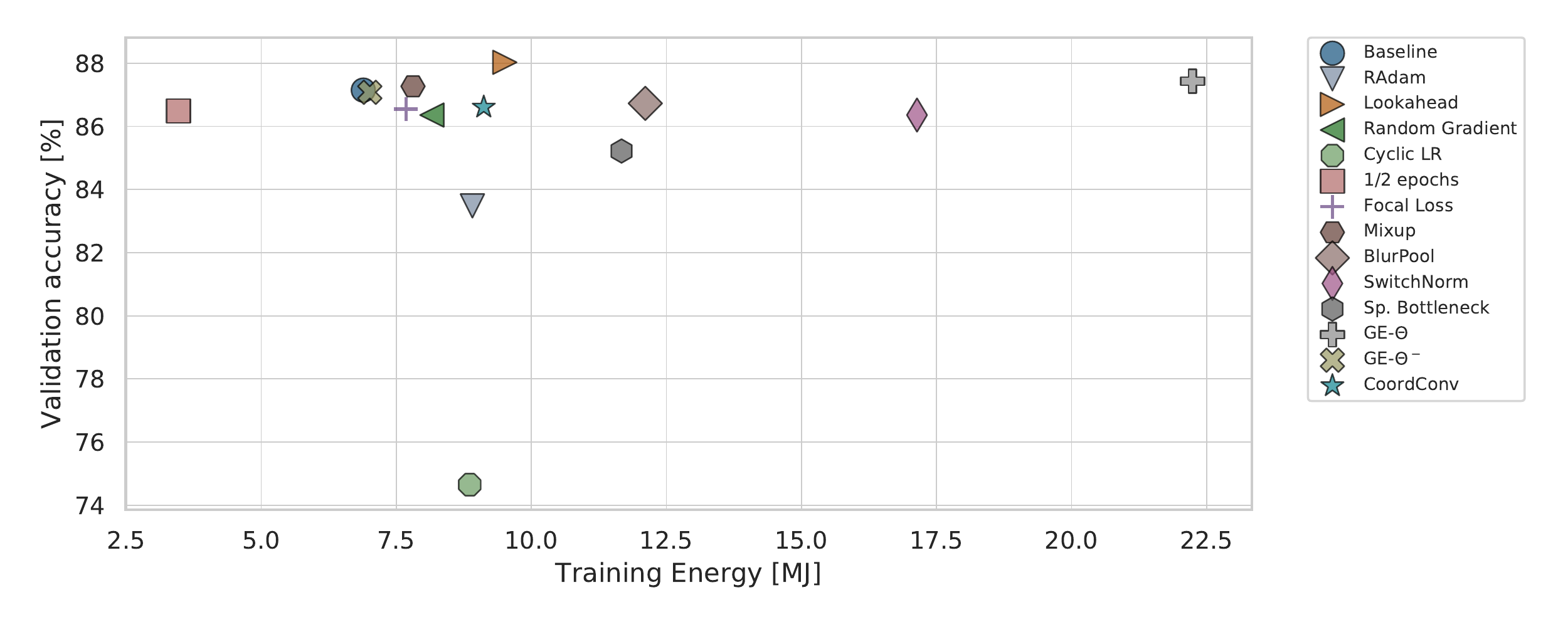}
\caption{Accuracy vs. training energy for each technique on \textbf{(a)} CIFAR-100 test set and \textbf{(b)} Tiny-ImageNet validation set : no correlation between the training energy and validation/test accuracy is observed.}
\label{fig:plot_validation_miou_vs_training_energy_cls}
\end{figure*}

\begin{figure*}[h]
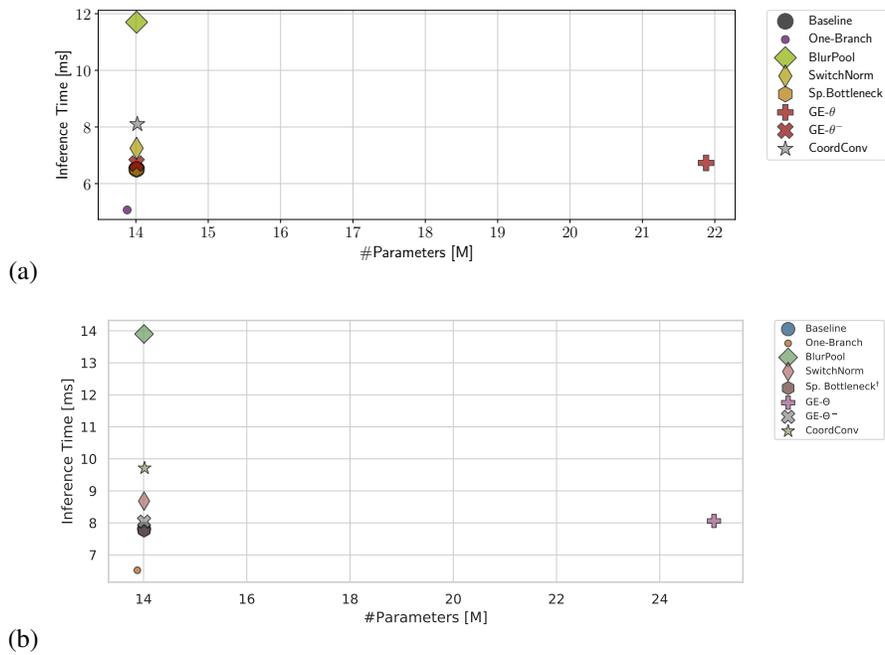

\centering
\subfigure(a){\includegraphics[width=0.7\linewidth]{./figures/plot_inference_time_vs_parameters.pdf}} \\
\subfigure(b){\includegraphics[width=0.7\linewidth]{./figures/plot_inference_time_vs_parameters_coco.pdf}}
\caption{Inference Time vs. Number of parameters for each technique on \textbf{(a)} CIFAR-100 test set and \textbf{(b)} Tiny-ImageNet validation set : The number of parameters is not the only determinant for inference time. This highlights the need to make effective use of parameters in NN design, for instance, ensuring minimal memory overhead, and high parallelization.}
\label{fig:plot_inference_time_vs_parameters_cls}
\end{figure*}

\begin{figure*}[h]
\centering
\subfigure(a){\includegraphics[width=0.7\linewidth]{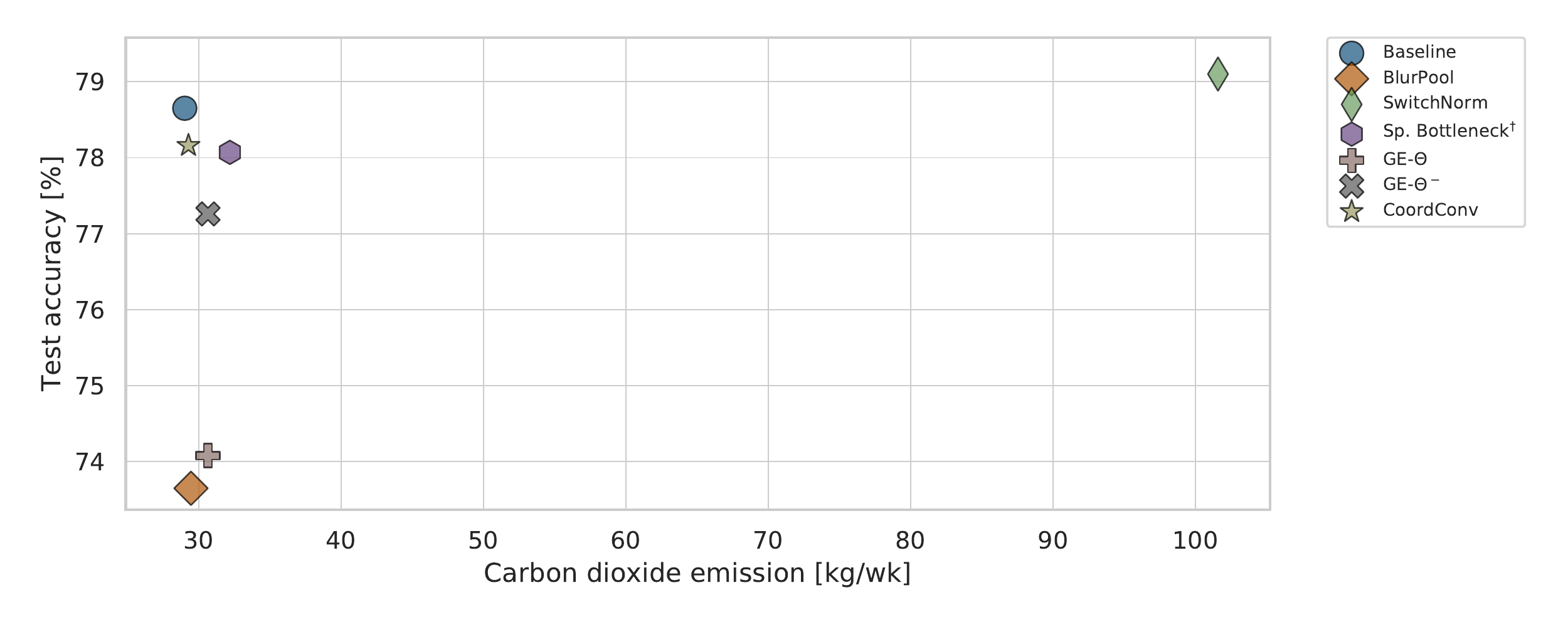}} \\
\subfigure(b){\includegraphics[width=0.7\linewidth]{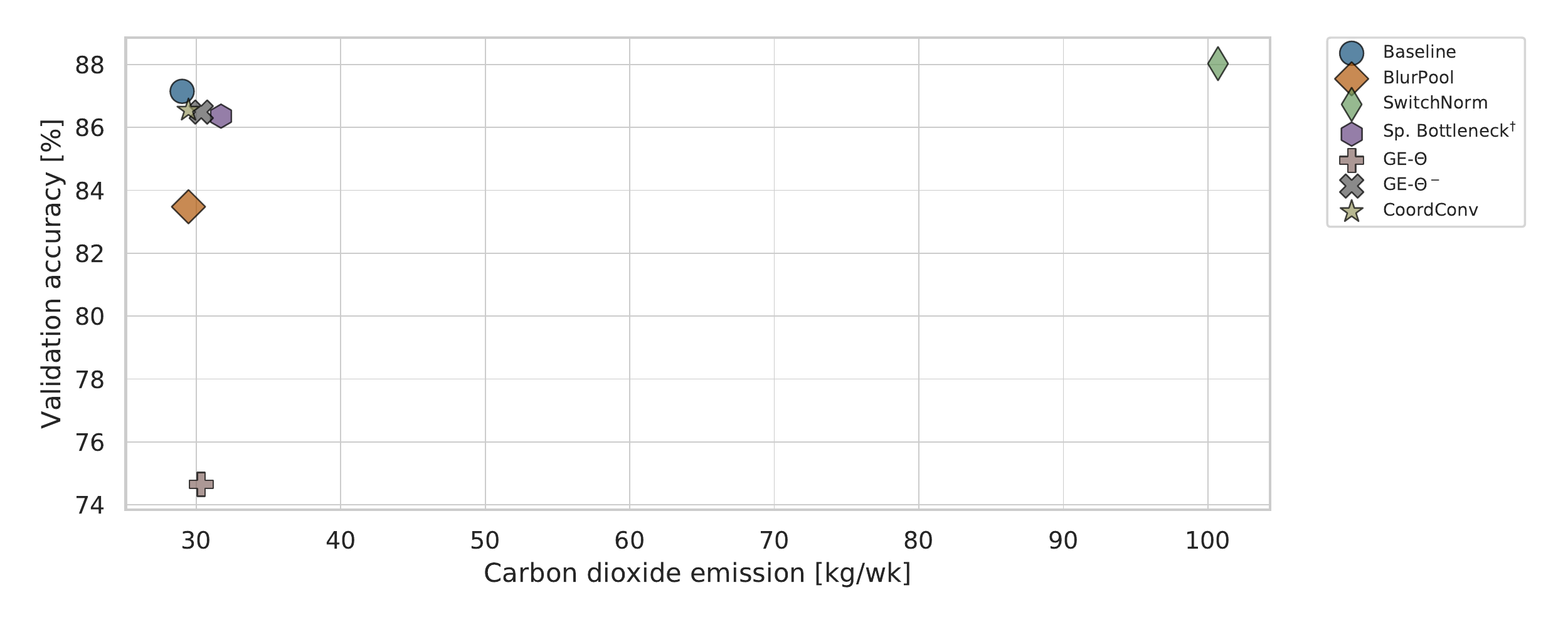}}
\caption{Accuracy vs. $CO_2$ emissions for each technique on \textbf{(a)} CIFAR-100 test set and \textbf{(b)} Tiny-ImageNet validation set : No correlation between test/val accuracy and $CO_2$ emission.}
\label{fig:plot_energy_vs_co2_cls}
\end{figure*}

\begin{figure*}[h]
\centering
\subfigure(a){\includegraphics[width=0.7\linewidth]{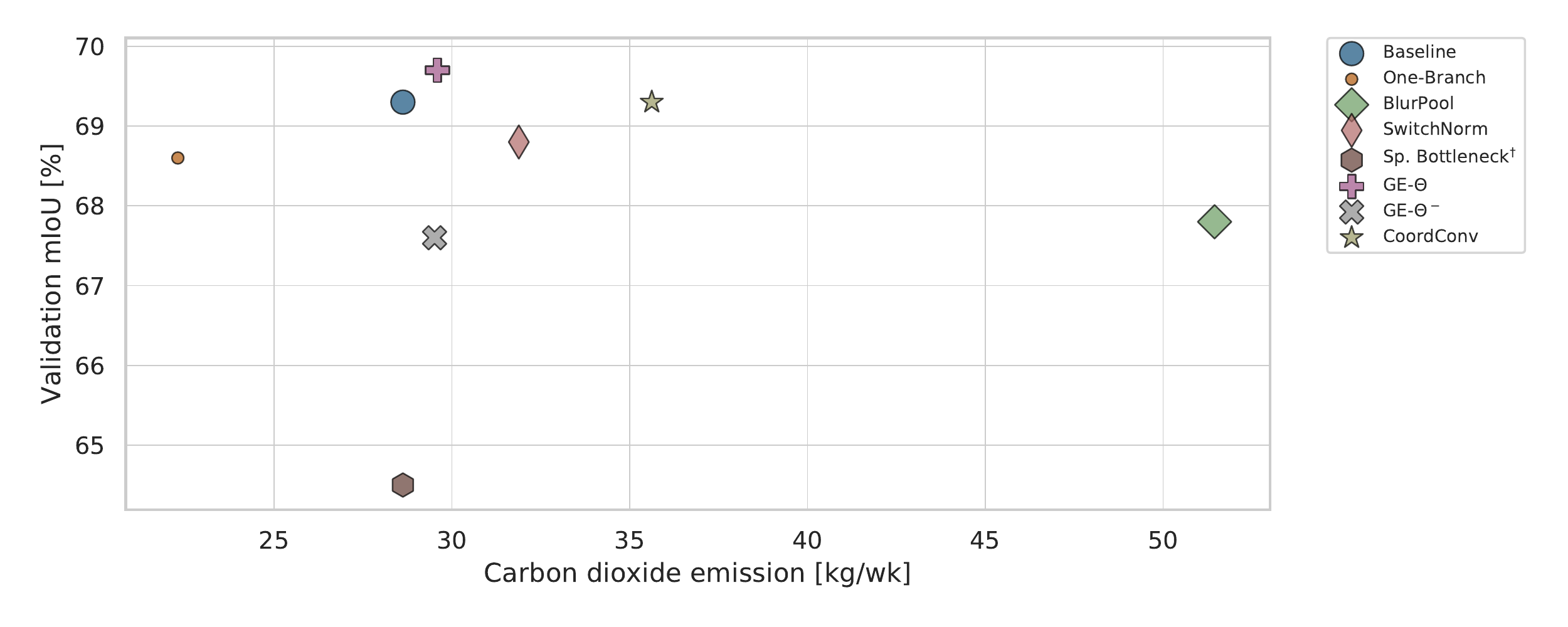}} \\
\subfigure(b){\includegraphics[width=0.7\linewidth]{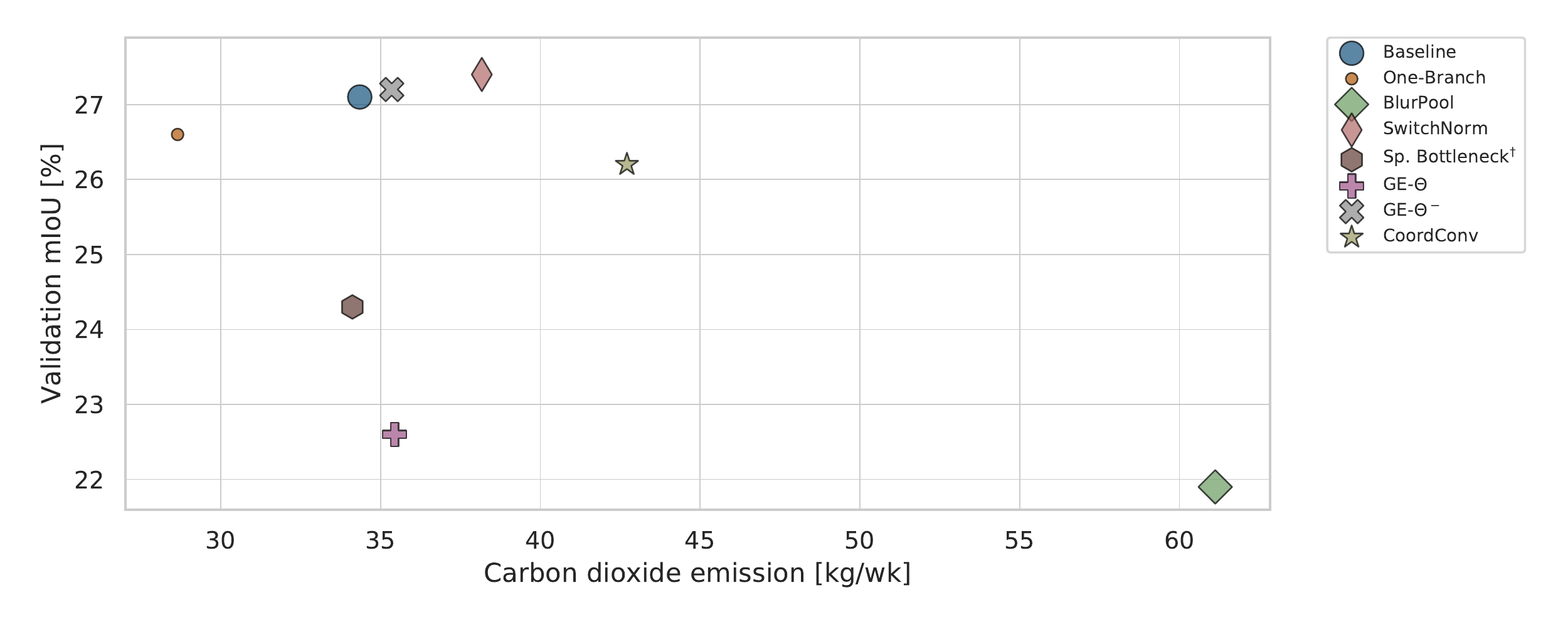}}
\caption{Validation mIoU vs. $CO_2$ emissions for each technique on \textbf{(a)} Cityscapes and \textbf{(b)} COCO-Stuff : No correlation between validation mIoU and $CO_2$ emission.}
\label{fig:plot_energy_vs_co2_seg}
\end{figure*}
\FloatBarrier

\end{document}